\documentclass[12pt]{article}

\usepackage{tikz}
\tikzset{font={\fontsize{8pt}{10}\selectfont}}

\usepackage{geometry}
\usetikzlibrary{calc}
\usetikzlibrary{arrows,shapes,chains}
\usepackage{xcolor}
\usepackage{amsfonts}	
\usepackage{amsmath}
\usepackage{bm}
\usepackage{graphicx}
\usepackage{multirow}
\usepackage{algorithmic}
\usepackage{algorithm}
\usepackage{subfigure}
\usepackage{epstopdf}
\usepackage{url}
\usepackage{cleveref}
\usepackage{cases}
\crefname{equation}{}{}
\crefname{table}{TABLE}{TABLES}
\crefname{figure}{Fig.}{Figs.}
\crefname{section}{Section}{Sections}

\newcommand{\cb}[1]{\ifmmode {\boldsymbol{#1}}\else ${\boldsymbol{#1}}$\fi}

\newcommand{\cp}[1]{\ifmmode {\mathcal{#1}}\else ${\mathcal{#1}}$\fi}

\newtheorem{thm}{Theorem}
\newtheorem{defination}{Defination}

\begin{document}

\title{Improving the Expressive Power of Graph Neural Network with Tinhofer Algorithm}
\author{Alan~J.X.~Guo,  Qing-Hu Hou, and Ou Wu}
\date{}

\maketitle

\begin{abstract}
    In recent years, Graph Neural Network (GNN) has bloomly progressed for its power in processing
    graph-based data. 
    Most GNNs follow a message passing scheme, 
    and their expressive power is mathematically limited by the discriminative ability of the Weisfeiler-Lehman (WL) test.
    Following Tinhofer's research on compact graphs, 
    we propose a variation of the message passing scheme, called 
    the Weisfeiler-Lehman-Tinhofer GNN (WLT-GNN), 
    that theoretically breaks through the limitation of the WL test. 
    In addition, we conduct comparative experiments and ablation studies on several well-known datasets. 
    The results show that the proposed methods have comparable performances and better expressive power on these datasets. 
\end{abstract}

\section{Introduction}
\label{introduction}
Graphs are the basic structures of a large amount of data analysis work, 
including social networks \cite{zweig2016graph}, biological networks \cite{gao2018study}, 
chemical networks \cite{trinajstic2018chemical}, {\em etc.}. 
Recently, the Graph Neural Network (GNN) \cite{gori2005new,scarselli2009graph} has gained much attention 
due to its ability to utilize information representing the structure of a graph \cite{wu2020comprehensive,zhang2020deep}.

Generally, typical GNN methods follow a scheme called message passing. 
In the message passing scheme, each node in the graph aggregates the information of its neighbors and 
then updates its own feature vector. 
After $k$ iterations of message passing, 
the entire graph is represented by reading the feature vectors of all nodes in the graph.
Different implementations of message passing and readout lead to different GNN algorithms.
In \cite{hamilton2017inductive}, the authors proposed GraphSAGE, which aggregates the neighbors' information 
by averaging feature vectors in the neighborhood. 
In \cite{kipf2017semi}, the authors proposed the graph convolution network (GCN), which is based on the first-order 
approximation of spectral convolutions on graphs. 
In GCN, the nodes aggregate the information by weighting its neighbors message. 
Attention mechanism \cite{vaswani2017attention} has also been introduced to message passing. 
The researchers of graph attention network (GAT) \cite{velickovic2018graph} use the feature vectors of nodes and their neighbors 
to query the weights of their neighbors during aggregating. 
Moreover, anothor work in \cite{zhang2020adaptive} added the structural 
fingerprint information while implementing the attention mechanism in GNN. 

However, the power of typical message passing scheme is bounded by the Weisfeiler-Lehman (WL) test
\cite{weisfeiler1968reduction}. 
This is because the WL test and the message passing mechanism share the same algorithm, 
but the WL test does not lose any information mathematically when performing aggregation. 
Passing the WL test is a necessary condition for a pair of graphs to be isomorphic; 
and the probability that WL test fails goes to $0$ when the size of the graph increases to infinity 
\cite{babai1980random}. 
However, there are still some graph classes that fail in the WL test and can not be ignored.
A simple example is that the WL test can not distinguish between $k$ regular graphs of the same size.

In order to reach and break through the limitation of the WL test in GNN, researchers have made several attempts.
Graph Isomorphism Network (GIN) in \cite{xu2018how} established the message passing with injection functions, 
and theoretically reached the limitation of WL test.
Higher order WL tests have also been introduced, 
such as $k$-WL, folklore $k$-WL, and set $k$-WL tests,
to gain the power of GNN. 
These tests are usually more powerful or at least not inferior than the WL test. 
The set $k$-WL algorithm based GNN is proposed in \cite{morris2019weisfeiler}, 
while the $k$-WL and folklore $k$-WL algorithms and their related theoretical researches 
are \cite{maron2019provably,maron2019invariant,maron2019universality,chen2019equivalence,chen2020can}. 
There are also some heuristic methods that try to break the limitaion of WL test. 
In \cite{xu2018representation} the authors enable structure-aware representations in their proposed 
jumping knowledge (JK) networks with different neighborhood ranges. 
In \cite{bouritsas2020improving}, the authors encode the neighborhood structures by counting graph substructures to 
gain the power of GNN. 
A comprehensive survey on the relations between WL test and GNN is provided by \cite{sato2020survey}.

In the work \cite{tinhofer1991note}, the authors proposed a WL-based algorithm, called Algorithm GRAPHIS in their paper, 
to study the isomorphism between compact graphs \cite{tinhofer1986graph}. 
By executing the Tinhofer algorithm, any pair of non-isomorphic graphs can be distinguished.
Moreover, if one of a pair of graphs is a compact graph, 
no matter what the other graph is, Tinhofer algorithm always gives the correct answer.
In this view, passing the Tinhofer test could be regarded as a fine-grained judgment of isomorphism. 

In this paper, by proposing a newly designed recoloring layer, 
we introduce the fine-grained algorithm from Tinhofer's mathematical work to 
GNN's message passing scheme. 
The proposed Weisfeiler-Lehman-Tinhofer GNN (WLT-GNN) theoretically break through the limitation of the WL test.
We also conduct experiments on well-known data sets, 
showing that the proposed WLT-GNN has comparable performance to the state of art on these datasets, 
and the introducing of a recoloring layer helps to improve the expressive ability of GNN.


\section{Notations and Preliminaries}
\label{preliminaries}
In this section, we introduce some notations of graph theory and GNN. 
We try our best to use unified notations for all the following contents, including the message passing scheme, 
the WL test, the Tinhofer test, and the proposed WLT-GNN. 

Let $G=(V(G),E(G);\bm{X}(G))$ be a graph, $V(G)$ be the set of nodes or vertices of $G$, $E(G)$ be the set 
of edges of $G$, and $\bm{X}(G)=(\bm{x}_1,\bm{x}_2\ldots,\bm{x}_n)$ be the node features of nodes $1,2,\ldots,n\in V(G)$. 
Given a finite graph $G$, 
we use a square $0$-$1$ matrix $\bm{A}$ to indicate whether the vertices in $G$ are adjacent or not , 
and call this matrix the adjacency matrix of $G$. 
Given two graphs without node features $G=(V(G),E(G))$ and $H=(V(H),E(H))$, 
we say that the graph $G$ is isomorphic to the graph $H$, iff we could 
find a bijection $\pi:V(G)\mapsto V(H)$ such that $(u,v)\in E(G)$ iff $(\pi(u),\pi(v))\in E(H)$.

In this paper, we consider the problem of graph classification. 
The graph classification predicts the label $y_G$ of 
a graph $G\in \mathcal{G}$ through function $f:G\mapsto y_G$, where $\mathcal{G}$ is a set of graphs. 
As a simple assumption, 
if nodes' feature is not considered, isomorphic graphs should be assigned with the same label.

Before focusing on the problem of graph classification, 
let us introduce the key part of a typical GNN, the message passing mechanism.
Let $v\in V(G)$ be a vertex of graph $G$, the vector $\bm{h}^{(t)}_v$ be the message of $v$ at time $t$. 
In message passing mechanism, the information flows by aggregating the messages from the neighbors of $v$. 
Let $f^{(t)}_{\mathrm{aggregate}}$ and $f^{(t)}_{\mathrm{update}}$ be the funtions used for aggregating and updating 
the messages at time $t$, the message $\bm{h}_v^{(t+1)}$ of $v$ at time $t+1$ could be formulated as:
\begin{eqnarray} 
    \bm{a}_v^{(t)} &=& f^{(t)}_{\mathrm{aggregate}}(\{\{\bm{h}^{(t)}_u|\, u\in \mathcal{N}(v)\}\});\label{message_passing_1}\\
    \bm{h}_v^{(t+1)} &=& f^{(t)}_{\mathrm{update}}(\bm{a}_v^{(t)},\bm{h}^{(t)}_v).\label{message_passing_2}
\end{eqnarray}
where
the notation $\{\{\ldots\}\}$ is used to denote multiset that allows repeated elements, 
and the $\mathcal{N}(v)$ is used to denote the set of neighbors of $v$ in $G$. 

In the practice of applying GNN for graph classification, each node $v$ of the graph is assigned with 
a initial message $\bm{h}^0_v$, usually the encoding of degree of $v$ or a constant number. 
After $k$ times iteration with Equations (\ref{message_passing_1}, \ref{message_passing_2}), 
a readout function $f_{\mathrm{readout}}$ is used to read all the messages into $\bm{h}_G$ from the graph
\begin{equation}
    \bm{h}_G = f_{\mathrm{readout}}(\{\{\bm{h}^{(k)}_v|\, v\in V(G)\}\}).
\end{equation}
Finally, the message $\bm{h}_G$ of the graph is used to predict the label $\hat{y}_G$. 

For example, the GraphSAGE \cite{hamilton2017inductive} used the following aggregation and update functions:
\begin{eqnarray}
    f^{(t)}_{\mathrm{aggregate}}(\{\{\bm{h}^{(t)}_u|\, u\in \mathcal{N}(v)\}\}) &=& \sum_{u\in\mathcal{N}(v)}\frac{\bm{h}^{(t)}_u}{\deg(v)};\label{graphsage-1}\\
    f^{(t)}_{\mathrm{update}}(\bm{a}_v^{(t)},\bm{h}^{(t)}_v) &=& \sigma(\bm{W}^{(t)}[\bm{a}_v^{(t)},\bm{h}^{(t)}_v]),
\end{eqnarray}
where the $\bm{W}^{(t)}$ is the linear transformation and the $\sigma$ is the activation function. 

\section{Weisfeiler-Lehman Test and Tinhofer Algorithm}
\label{WLT-math}
In this section, we mainly introduce the WL test and the Tinhofer algorithm. 
We also conduct some theoretical analysis related to GNN on these algorithms and message passing schemes.
Because the WL algorithm uses the word ``color'' to represent the vertex message, 
we will not distinguish between the words ``color'' and ``message'', they both represent 
a node's message $\bm{h}_u$. 

The WL test \cite{weisfeiler1968reduction} is a fast algorithm based on vertex color refinement 
for the graph isomorphism problem; 
it gives answers of ``non-isomorphic'' and ``possible isomorphic'' on a pair of graphs. 
Using the notations of message passing, the WL algorithm can be expressed as Algorithm~\ref{wl-test}. 
In this algorithm, the aggregation and update functions in message passing are implemented by the HASH function, 
which is an injection function that maps different inputs to different outputs. 
If we use color refinement to discribe the WL test, all nodes are first colored with $\bm{0}$ at time $t=0$. 
After that, during each iteration, each node $v$ is assigned with a new color 
$\bm{h}^{(t+1)}_v = \mathrm{HASH}(\{\{\bm{h}^{(t)}_u|\, u\in \mathcal{N}(v)\}\})$ 
that is uniquely calculated based on the colors of its neighbors. 
When the multisets of node colors of the two graphs $G,H$ are different, the test outputs ``non-isomorphic'' and exits. 
Or, when the color distribution of the nodes is stable and there is no ``non-isomorphic'' answer, 
the test outputs ``possible isomorphic'' on $G,H$. 
The stopping criterion of ``convergence'' is that no further refinement of 
$\{\{\bm{h}^{(t)}_u|\, u\in \mathcal{N}(v)\}\}$ is achieved at time $t+1$. 
It is theoretically ensured that the Algorithm~\ref{wl-test} stops after at most $|V(G)|+|V(H)|$ iterations \cite{cai1992optimal}. 

\begin{algorithm}[tb]
    \caption{Weisfeiler-Lehman Algorithm}
    \label{wl-test}
    \begin{algorithmic}
        \STATE {\bfseries Input:} A pair of graphs $G=(V(G),E(G))$, $H=(V(H),E(H))$. 
        \STATE Initialization: $\bm{h}^{(0)}_v\leftarrow \bm{0},\,\forall v\in V(G)$; $\bm{h}^{(0)}_u\leftarrow \bm{0},\,\forall u\in V(H)$; 
        $t\leftarrow 0$.
        \REPEAT
        \IF{$\{\{\bm{h}^{(t)}_v|\,\forall v\in V(G)\}\} \neq \{\{\bm{h}^{(t)}_u|\,\forall u\in V(H)\}\}$}
        \STATE return ``non-isomorphic''.
        \ENDIF
        \STATE $\bm{h}^{(t+1)}_v = \mathrm{HASH}(\{\{\bm{h}^{(t)}_u|\, u\in \mathcal{N}(v)\}\}), \forall v \in V(G);$
        \STATE $\bm{h}^{(t+1)}_u = \mathrm{HASH}(\{\{\bm{h}^{(t)}_u|\, u\in \mathcal{N}(v)\}\}), \forall v \in V(H);$
        \STATE $t=t+1;$
        \UNTIL{``convergence'';}
        \STATE return ``possible isomorphic''.
    \end{algorithmic}
\end{algorithm}

It's trival that passing the WL test is a necessary condition to make a pair of graphs isomorphic. 
In addition, when the order of the graph goes to infinity, the fraction of the non-isomorphic graphs that 
passes the WL test goes zero \cite{babai1980random}. 
However, the set of graphs that failed the WL test include important and meaningful graphs from the real world. 
For example, the WL test can not distinguish regular graphs even if they have different connected components. 
Further, suppose we have a pair of graphs $G,H$ and their adjacency matrices $\bm{A},\bm{B}$, respectively, 
the pair of graphs $G,H$ pass the WL test is equivalent to that 
the following linear program (\ref{lp1}, \ref{lp2}, \ref{lp3}) has feasible solution
\cite{tinhofer1991note,grohe2017descriptive}
\begin{eqnarray}
        \bm{X}\bm{A} &=& \bm{B}\bm{X}; \label{lp1}\\
        \bm{X}\bm{e} &=& \bm{X}^t\bm{e} = \bm{e}; \label{lp2}\\
        \bm{X} &\geq& 0. \label{lp3}
\end{eqnarray}
The $\bm{e}$ in Equation (\ref{lp2}) represents a vector filled with $1$s.
The Equation (\ref{lp2}) restricts the matrix $\bm{X}$ to a doubly stochastic matrix, 
the sums of whose rows and columns are $1$s. 
A permutation matrix $\bm{P}$ is a special doubly stochastic matrix, with only one $1$ per row and per column. 
In the linear program (\ref{lp1}, \ref{lp2}, \ref{lp3}),
replace the restriction of the doubly stochastic matrix $\bm{X}$ by the permutation matrix $\bm{P}$, 
the new linear program is solvable is equavilant to graphs $G,H$ are isomorphic.
The permutation $\pi$ defined by 
the permutation matrix $\bm{P}$ is an isomorphic map between $V(G)$ and $V(H)$. 
Therefore, the margin between passing WL test and isomorphism is the ``same'' with the difference between 
the two linear programs with 
the doubly stochastic matrix $\bm{X}$ and the permutation matrix $\bm{P}$, respectively. 

Typical GNNs are not reaching the power of WL test in distinguishing non-isomorphic graphs. 
Taking GraphSAGE as an example, in Equation (\ref{graphsage-1}), the aggregation function averages the messages of neighbors; 
this aggregation function is obivouse not an injection, so theoretically less powerful than the $\mathrm{HASH}$ 
function in WL algorithm. 
The authors of GIN \cite{xu2018how} used multi-layer perceptrons (MLP) on a summing collection 
of the neighbors messages as aggregation function, and mathematically proved that GIN is as powerful as WL test 
by the universal approximation theorem \cite{hornik1989multilayer}. 

In order to study the isomorphism between compact graphs, Tinhofer proposed the 
algorithm GRAPHIS (Tinhofer algorithm) in \cite{tinhofer1991note}, 
which works not only on compact graphs, but also on all the graphs. 
Before introducing the Tinhofer algorithm, let's define some notations. 
Using $\{V_1,V_2,\ldots,V_k\}$ to denote the collection of non-empty and disjoint subsets of $V$, if 
the union of these subsets is $V$, we call it a set partition of $V$. 
Gathering the same colored nodes, we use 
\begin{equation}
    \mathcal{V}^{(t)}(G) = \{V_1^{(t)}(G),\ldots,V_{k(t)}^{(t)}(G)\}, 
\end{equation} 
to denote the color partition of $V(G)$ at iteration $t$, 
where nodes belonging to the same subset have the same color, and the total number of colors is $k(t)$. 
In the WL test, we could rewrite the multiset of node colors at iteration $t$ by 
\begin{eqnarray}
    \mathcal{C}^{(t)}(G) &=& \{\{\bm{h}^{(t)}_v|\,\forall v\in V(G)\}\} \notag\\
    &=&\{(\bm{h}^{(t)}_1,V_1^{(t)}(G)),\ldots,(\bm{h}^{(t)}_{k(t)},V_{k(t)}^{(t)}(G))\}. 
\end{eqnarray}
Instead of coloring all the nodes in $V(G)$ with $\bm{0}$ in the Algorithm \ref{wl-test}, 
if we initiate the node colors with $\mathcal{C}=\{(\bm{h}_1,V_1(G)),\ldots,(\bm{h}_{k},V_{k}(G))\}$, 
the WL algorithm can still reach convergence. 
We use the ``closure'' of WL algorithm on $\mathcal{C}$ to call the converged multiset of node colors, 
denoted by $\mathrm{CLOSURE}_G(\mathcal{C})$. 
For example, the multiset of converged node colors is 
$\mathrm{CLOSURE}_G(\{(\bm{0},V(G))\})$, in the Algorithm \ref{wl-test}. 
With these notations, the Tinhofer algorithm is descriped in Algorithm \ref{tinhofer-algorithm}. 

As shown in Algorithm \ref{tinhofer-algorithm}, the Tinhofer algorithm is based on the closure of the WL algorithm. 
During each iteration, firstly, the converged node colors are computed by the WL algorithm with initial node colors; 
secondly, the converged state of node colors produced by WL algorithm is interrupted with a heuristic recoloring operation; 
finally, the recolored graph is considered as the input of the next iteration. 
The algorithm stops until the WL algorithm gives ``non-isomorphic'' answer or there is only one node 
in each subset of the converged color partition of $V(G)$.

Instead of the WL algorithm that always give correct answers on a pair of isomorphic graphs, the Tinhofer algorithm 
always give correct answers on a pair of non-isomorphic graphs \cite{tinhofer1991note}. 
However, this does not support the power of Tinhofer algorithm in graph isomorphism problems, 
because an algorithm that always say ``non-isomorphic'' to any pair of graphs also gives the correct answer on 
a pair of non-isomorphic graphs. 
Theorem~\ref{thm:compact} guarantees the Tinhofer algorithm's correctness on some classes of isomorphic graph pairs. 
\begin{thm}[\cite{tinhofer1991note}]\label{thm:compact}
    If $G$ is a compact graph, then each run of Algorithm \ref{tinhofer-algorithm} applied to $G$ and an arbitrary graph $H$ of 
    the same order as $G$ decides correctly whether $G$ is isomorphic to $H$ or not.
\end{thm}
The proof of Theorem~\ref{thm:compact} can be found in their original paper \cite{tinhofer1991note}, and 
a brief introduction of 
compact graph \cite{tinhofer1986graph} can be found in the Appendix.
It is worth noting that some graphs that can not be identified by WL test are compact, 
and therefore can be identified by Tinhofer algorithm. 
An example of compact regular graphs could be found in \cite{wang2005compact}. 
As mentioned above, a compact regular graph is regular and therefore fails WL test. 
However, 
it is also a compact graph that could be identified by Tinhofer algorithm. 
Moreover, it is proved in \cite{arvind2017graph} that if WL test could distinguish a graph $G$ from 
any non-isomorphic graph $H$, then the graph $G$ is compact. 

\begin{algorithm}[tb]
    \caption{Tinhofer Algorithm}
    \label{tinhofer-algorithm}
    \begin{algorithmic}
        \STATE {\bfseries Input:} A pair of graphs $G=(V(G),E(G))$, $H=(V(H),E(H))$ with $|V(G)| = |V(H)|$. 
        \STATE Initialization: $G$ with nodes' color $\mathcal{C}_G = \{(\bm{0},V(G))\}$; 
        $H$ with nodes' color $\mathcal{C}_H = \{(\bm{0},V(H))\}$. 
        \REPEAT
        \STATE Run WL algorithm on $G$ with $\mathcal{C}_G$ and $H$ with $\mathcal{C}_H$, 
        get 
        {\footnotesize 
        \begin{eqnarray*}
            &\mathrm{CLOSURE}_G(\mathcal{C}_G) =\{(\bm{h}_1,V_1(G)),\ldots,(\bm{h}_{k},V_{k}(G))\};\\
            &\mathrm{CLOSURE}_H(\mathcal{C}_H) =\{(\bm{h}_1,V_1(H)),\ldots,(\bm{h}_{k},V_{k}(H))\};
        \end{eqnarray*}}
        \IF{$\mathrm{CLOSURE}_G(\mathcal{C}_G) \neq \mathrm{CLOSURE}_H(\mathcal{C}_H)$}
        \STATE Return ``possible non-isomorphic''.
        \ENDIF
        \IF{$\mathrm{len}(\mathrm{CLOSURE}_G(\mathcal{C}_G))=k < |V(G)|$}
        \STATE {\textsl{\# The recoloring procedure. }}
        \STATE Choose $i$ such that $|V_i(G)|>1$; 
        \STATE Choose nodes $v\in V_i(G), u\in V_i(H)$;
        \STATE Recolor $v,u$ with a new color $\bm{h}_{k+1} = \mathrm{HASH}(\bm{h}_{i})$, and update $\mathcal{C}_G,\mathcal{C}_H$:
        {\footnotesize 
        \begin{eqnarray*}
            \mathcal{C}_G=&\{(\bm{h}_1,V_1(G)),\ldots,(\bm{h}_{i},V_{i}(G)\backslash\{v\}),\\
            &\ldots,(\bm{h}_{k},V_{k}(G)),(\bm{h}_{k+1},\{v\})\};\\
            \mathcal{C}_H=&\{(\bm{h}_1,V_1(H)),\ldots,(\bm{h}_{i},V_{i}(H)\backslash\{u\}),\\
            &\ldots,(\bm{h}_{k},V_{k}(H)),(\bm{h}_{k+1},\{u\})\};
        \end{eqnarray*}
        }
        \ENDIF
        \UNTIL{$\mathrm{len}(\mathrm{CLOSURE}_G(\mathcal{C}_G))=k = |V(G)|$;}
        \STATE return ``isomorphic''.
    \end{algorithmic}
\end{algorithm}

As shown in this section, the WL test gives coarse-grained answers to isomorphism, while the Tinhofer test gives 
fine-grained answers. 
Moreover, the Tinhofer algorithm is based on the WL test. 
Through the first iteration of Algorithm \ref{tinhofer-algorithm}, 
the power of the WL test can be easily expressed in the Tinhofer algorithm.
In theory, we are able to classify different fine-grained information into a unified class, 
but we can not divide the coarse-grained superclass into several subclasses without more information. 
Since typical GNNs are limited by the power of WL algorithm for its message passing scheme, 
a straightforward idea is to construct a type of GNN to simulate the Tinhofer test and break through the limitation 
of the WL test. 

\section{Proposed Weisfeiler-Lehman-Tinhofer GNN}
\label{WLT-GNN}
In this section, we give a detailed description of the proposed WLT-GNN. 
It can be seen that the Tinhofer algorithm is mainly composed of the WL algorithm and the recoloring procedure.
The WLT-GNN we proposed is also composed of two corresponding layers, namely the GIN layer and the recoloring layer.

In the work of \cite{xu2018how}, the authors proposed the GIN and provided mathematical 
proof that GIN can reach the power of WL test. 
In our work, we use the GIN layer to simulate the WL algorithm 
in Algorithm \ref{tinhofer-algorithm}.
Under the GIN architecture, the messages of iteration $t+1$ is calculated on the 
messages of iteration $t$ by the following equation, 
\begin{equation}\label{eq:gin}
    \bm{h}_v^{(t+1)} = 
    \mathrm{MLP}^{(t)}\left(
        \left(1+\epsilon^{(t)}\right) \bm{h}_v^{(t)}
        + \sum_{u\in \mathcal{N}(v)}\bm{h}_u^{(t)}
    \right),
\end{equation}
where the $\mathrm{MLP}^{(t)}$ is the update function of iteration $t$ 
that fulfilled by a $2$-layered MLP. 
According to the $\epsilon$ is constant $\epsilon=0$ or trainable in the Equation (\ref{eq:gin}), 
GIN has two variations, which are expressed as GIN-$0$ and GIN-$\epsilon$, respectively. 

The recoloring layer is the main contribution of this article. 
It simulates the recoloring procedure of Algorithm \ref{tinhofer-algorithm}.
Suppose that the messages of the nodes $V(G)$ of the graph $G$ produced by the previous 
GNN layer is 
\begin{equation}
    \{\{\bm{h}_v|\,\forall v\in V(G)\}\},
\end{equation}
which can also be rewritten in the set partition format as 
\begin{equation}
    \{(\bm{h}_1,V_1(G)),\ldots,(\bm{h}_{k},V_{k}(G))\}, 
\end{equation}
where $k$ is asserted to be less than $|V(G)|$. 
In Algorithm \ref{tinhofer-algorithm}, the recoloring procedure is applied on a pair of graphs $(G,H)$ 
by choosing and recoloring a pair of nodes $(v,u)$ from the $i$-th subsets $(V_i(G),V_i(H))$ of the color partitions 
$(\mathrm{CLOSURE}_G(\mathcal{C}(G)),\mathrm{CLOSURE}_H(\mathcal{C}(H)))$, respectively. 
However, GNN is not designed to distinguish a pair of non-isomorphic graphs as the WL or Tinhofer algorithms do. 
It processes a single graph $G$ at each run, and predicts the label of $G$ based on the its output features. 
Therefore, when we apply the recoloring procedure in our proposed WLT-GNN, 
we can not randomly select $V_i(G)$ as the recoloring candidate set, 
but need to ensure that the same $V_i(G)$ is selected in different runs on the same $G$ or $G$'s isomorphic graphs. 
To meet this requirement, we sort the vectors 
\begin{equation}
    \tilde{\bm{h}}_i = \mathrm{concat}([|V_i(G)|],\bm{h}_i), 
\end{equation}
which are formed by concatenating the number of nodes in $V_i(G)$ and the message $\bm{h}_i$, 
and pick the largest $\tilde{\bm{h}}_{i_0}$ under the lexicographic order. 
With the chosen $\tilde{\bm{h}}_{i_0}$ and ${i_0}$, the recolored node $v$ is randomly 
choosed from $V_{i_0}(G)$. 
It can be easily verified that the randomness of choosing $v$ in $V_{i_0}(G)$ will not violate the aforementioned requirement. 
Finally, to recolor the chozen node $v$, 
we replace the message of $v$ from $\bm{h}_{i_0}$ to $\bm{0}$.

If we use letter $\mathrm{g}$ to represent the GIN layer and letter $\mathrm{r}$ to represent the recoloring layer, 
a typical structure of WLT-GNN is to apply GIN layers and recoloring layers sequentially, 
for example, $\mathrm{gggrgg}$ 
means stacking three GIN layers, one recoloring layers, and two GIN layers. 
In order to perform further classification tasks, people usually use MLP to classify the features 
globally readout on the last layer of the GNN. 
However, the output of the last layer of the proposed WLT-GNN does not explicitly express the features related to the 
WL algorithm in the Tinhofer algorithm, 
therefore, we use the jumping knowledge (JK) \cite{xu2018representation} strategy to 
collect the features produced by each layer $\mathrm{g}$. 
To be precise, we collect the global readouts of $\mathrm{g}$ for each layer 
and apply a weighted sum to these readouts. 
Finally, we use ordianry MLP and Softmax functions on the features and predict the labels of input graphs. 

The Tinhofer's proof supports one node recoloring in each iteration. 
However, when dealing with large scaled graphs, 
the recoloring of one node in $V_{i_0}(G)$ may be like a drop of ink in the ocean. 
To avoid this potential issue, we heuristically try to increase the number of 
recolored nodes in each iteration. 
In practice, we randomly recolor half of the nodes from $V_{i_0}(G)$ in the recoloring layer. 
This variant of WLT-GNN is denoted as WLT-GNN($0.5$) in the following text. 

\section{Experiments}
\label{experiments}
In order to show the effect of introducing Tinhofer algorithm to GNN, 
we conduct comparative experiments on the proposed WLT-GNN and some other well-known GNN structures. 
In addition, in order to indicate that the introducing of recoloring layer will improve the expressive ability of GNN, 
we also conduct ablation studies. 

Thanks to \textsl{PyTorch Geometric}
\footnote{https://github.com/rusty1s/pytorch\_geometric} 
\cite{fey2019fast}
and \textsl{TUDataset}\footnote{https://github.com/chrsmrrs/tudataset} \cite{morris2020tudataset}, 
they collected and implemented almost all relevant datasets and GNN structures in the same environment. 
They also reported the results on common models and datasets in \cite{fey2019fast}. 
In our paper, we use \textsl{PyTorch Geometric} for all the experiments, 
although the results may be different from the official reports. 

We engage seven commonly used datasets with more than $1000$ nodes for the comparative experiments. 
They are two bioinformatic datasets \cite{borgwardt2005protein,wale2006comparison}: PROTEINS and NCI1, 
and five social network datasets \cite{yanardag2015deep}: COLLAB, IMDB-BINARY, IMDB-MULTI, REDDIT-BINARY and REDDIT-MULTI-5K. 
For ablation studies, we use three representative datasets, namely PROTEINS, NCI1, and REDDIT-BINARY. 
If the dataset have node labels, they are used as the initial messages; 
otherwise, the one-hot encodings of the node degrees are used as the initial messages. 
The details of these datasets can be found in the Appendix.

\subsection{Comparative Experiments and Testing Performance}
The comparative experiments are conducted between the proposed WLT-GNN, WLT-GNN($0.5$) 
and the well-known GCN, GraphSAGE, GIN-$0$ and GIN-$\epsilon$. 
The comparative methods follow the settings in \cite{fey2019fast}. 
They use global mean operator and JK strategy to obtain features for further classification. 
The number of hidden units ($\in\{16,32,64,128\}$) and the number of layers ($\in\{2,3,4,5\}$) are tuned with 
respect to the validation set. 
The final result is reported by an average accuracy of $10$-fold cross validation, where the validation set 
is randomly choosed by $1$ fold from the $9$ training folds. 
For the proposed methods, we use fixed structure $\mathrm{gggrgg}$, which is stacking three GIN-$0$ layers, one recoloring 
layer, and two GIN-$0$ layers. 
We use global add operator to obtain the global readouts from the graph. 
The number of hidden units in the participating GIN-$0$ layers is $32$ for PROTEINS and $128$ for other datasets. 
The result is also reported by an average accruracy of $10$-fold cross validation. 
For fair comparison, the $8$ of the $9$ training folds are used 
for training, although the WLT-GNN and WLT-GNN($0.5$) have no hyperparameters need to be tuned by validation set. 
All the experiments are trained under the optimizer \textsl{Adam} \cite{kingma2015adam}
with $100$ epochs, in which the learning rate is $0.01$ and decays by multiplying $0.5$ at epoch $50$.

\begin{table}
    \caption{\textbf{Testing accuracies ($\%$).} 
        The accuracies are averaged over $10$-fold cross validation 
        and reported in the format $\mathrm{mean}\pm\mathrm{std}$. 
        The top-$2$ accuracies are highlighted with boldface.}
    \label{test_result}
    \begin{center}
    \begin{small}
    \begin{sc}
    \resizebox{\textwidth}{15mm}{
    \begin{tabular}{cccccccc}
    Method         & PROTEINS       & COLLAB         & IMDB-B         & IMDB-M & REDDIT-B       & REDDIT-M5K &NCI1\\
    GCN            & 73.1 $\pm$ 3.8        &\textbf{80.6 $\pm$ 2.1}&72.6 $\pm$ 4.5         &49.9 $\pm$ 3.4         &89.3 $\pm$ 3.3         &54.3 $\pm$ 1.6         &71.8 $\pm$ 3.6\\
    SAGE           & 73.8 $\pm$ 3.6        &79.7 $\pm$ 1.7         &72.4 $\pm$ 3.6         &49.5 $\pm$ 2.7         &89.1 $\pm$ 1.9         &52.7 $\pm$ 2.3         &74.5 $\pm$ 2.7\\
    GIN-$0$        & 72.1 $\pm$ 5.1        &79.3 $\pm$ 2.7         &72.8 $\pm$ 4.5         &49.7 $\pm$ 2.0         &89.6 $\pm$ 2.6         &55.7 $\pm$ 2.2         &75.7 $\pm$ 1.9\\
    GIN-$\epsilon$ & 72.6 $\pm$ 4.9        &79.8 $\pm$ 2.4         &72.1 $\pm$ 5.1         &48.4 $\pm$ 2.7         &90.3 $\pm$ 3.0         &\textbf{56.5 $\pm$ 1.8}&77.3 $\pm$ 1.5\\
    WLT-GNN        &\textbf{75.4 $\pm$ 3.7}&\textbf{80.2 $\pm$ 1.3}&\textbf{74.4 $\pm$ 6.4}&\textbf{51.3 $\pm$ 2.4}&\textbf{90.8 $\pm$ 1.7}&56.4 $\pm$ 1.8         &\textbf{77.8 $\pm$ 2.4}\\
    WLT-GNN($0.5$) &\textbf{74.8 $\pm$ 2.9}&80.0 $\pm$ 1.7         &\textbf{72.9 $\pm$ 4.3}&\textbf{51.4 $\pm$ 3.3}&\textbf{91.6 $\pm$ 0.9}&\textbf{56.8 $\pm$ 1.5}&\textbf{78.5 $\pm$ 2.6}\\
    \end{tabular}}
    \end{sc}
    \end{small}
    \end{center}
    \vskip -0.1in
\end{table}

The testing results are reported in Table~\ref{test_result}. 
It can be seen that, the proposed WLT-GNN and WLT-GNN($0.5$) outperform the GCN, GraphSAGE and GIN with a large 
margin on PROTEINS, IMDB-B, IMDB-M, REDDIT-B, and NCI1. 
On the dataset COLLAB and REDDIT-MULTI-5K, the proposed methods also show comparable results. 

\subsection{Ablation Studies and Expressive Power}

Since the testing performance highly depends on the model abilities of expression and generalization, 
the testing accuracy is a comprehensive metric for evaluating the model performances. 
However, the training accuracy is more related to the expressive power of the model. 
When evaluating expressive power, it is no longer necessary to consider generalization ability, 
and overfitting is also no longer an issue, because the model can not exceed its expressive ability and overfit on 
unknown information. 

In this paper, we conduct ablation studies by removing the recoloring layer 
in WLT-GNN and WLT-GNN($0.5$), without any other modifications. 
The experiments is performed on datasets PROTEINS, NCI1, and REDDIT-BINARY. 
Without concerning the overfitting issues, all the WLT-GNN and WLT-GNN($0.5$) are equipped with 
$128$ hidden units in their GIN-$0$ layers. 
By removing the recoloring layer, the ablation study of WLT-GNN only leaves the GIN-$0$ layers, 
which is a $\mathrm{ggggg}$ structured WLT-GNN. 
We denote this settings with GIN-$0$ in our study. 
All the models are trained for $300$ epochs under the optimizer \textsl{Adam}. 
The learning rate starts at $0.01$ and decays every $50$ epochs with multiplying $\sqrt{0.1}$. 
To illustrate the best expressive power of the conducted methods, 
the best performance of five runs is reported. 
Because we only considered training performance, the entire dataset is used as training data and 
there is no validation and testing sets. 

\begin{figure*}[ht]
    \vskip 0.2in
    \begin{center}
    \centerline{
        \subfigure[PROTEINS]{
            \begin{minipage}[b]{0.27\textwidth}
                \includegraphics[width=1\textwidth,trim=0 0 30 30,clip]{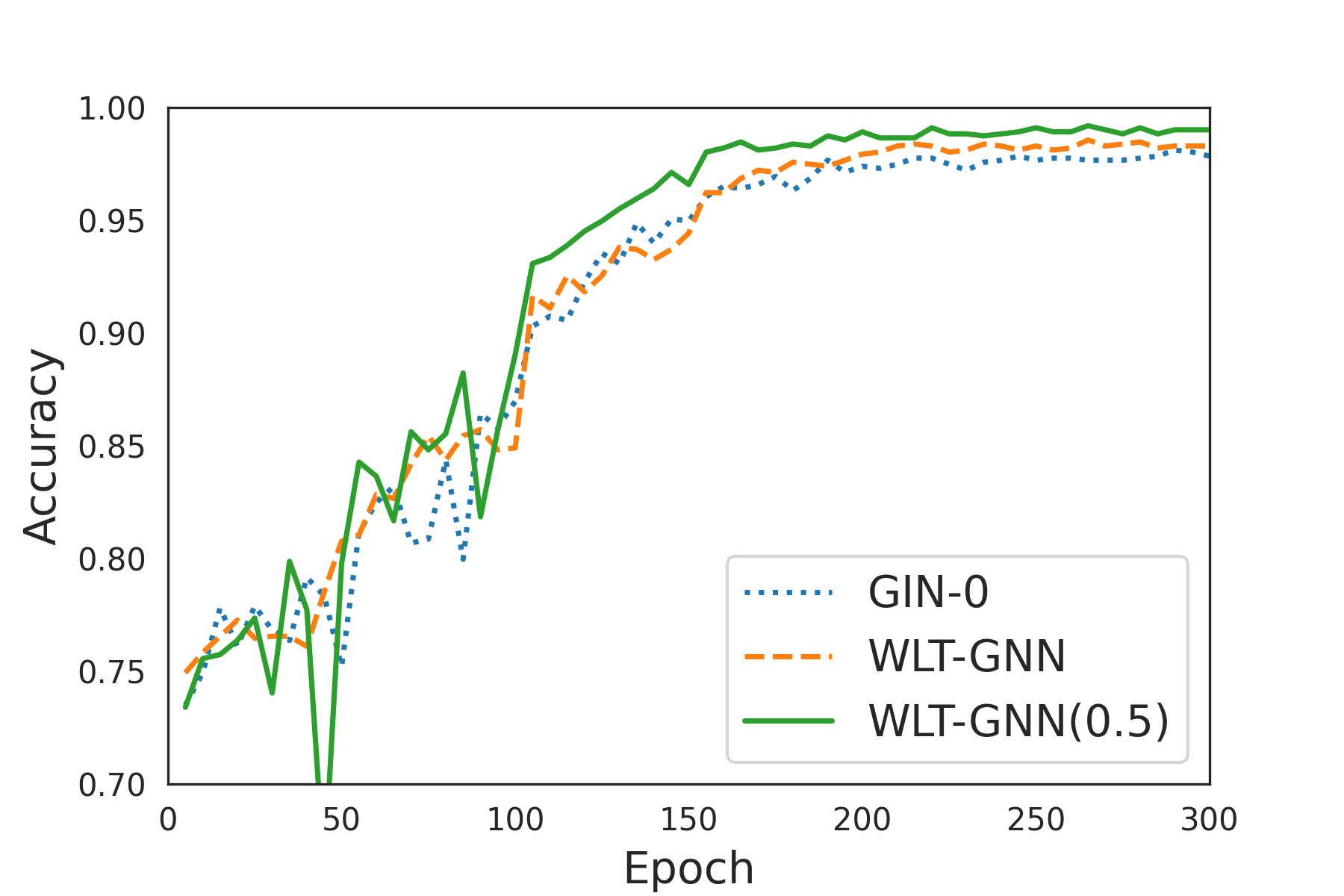}\\
                \includegraphics[width=1\textwidth,trim=0 0 30 30,clip]{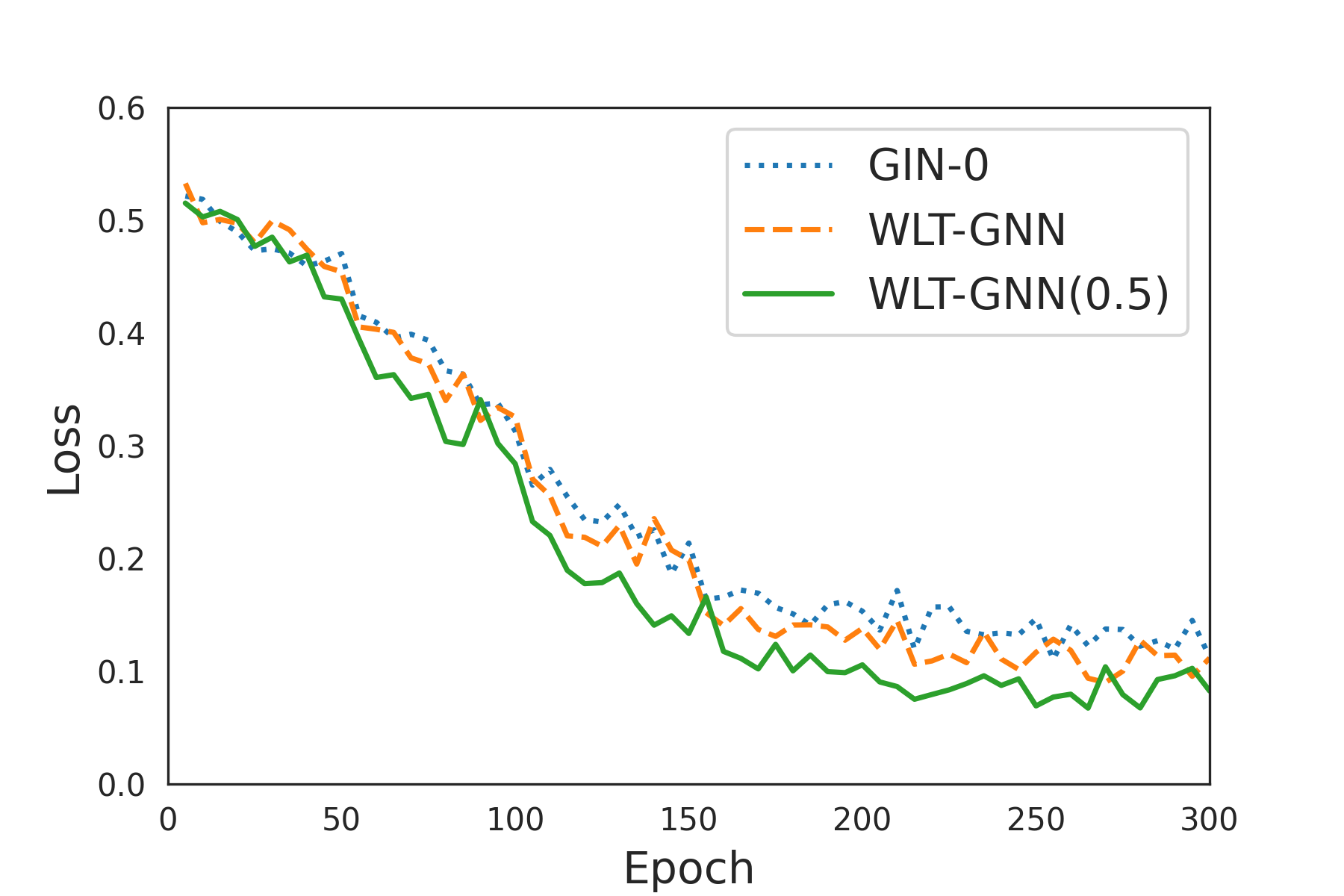}
            \end{minipage}
        }
        \subfigure[NCI1]{
            \begin{minipage}[b]{0.27\textwidth}
                \includegraphics[width=1\textwidth,trim=0 0 30 30,clip]{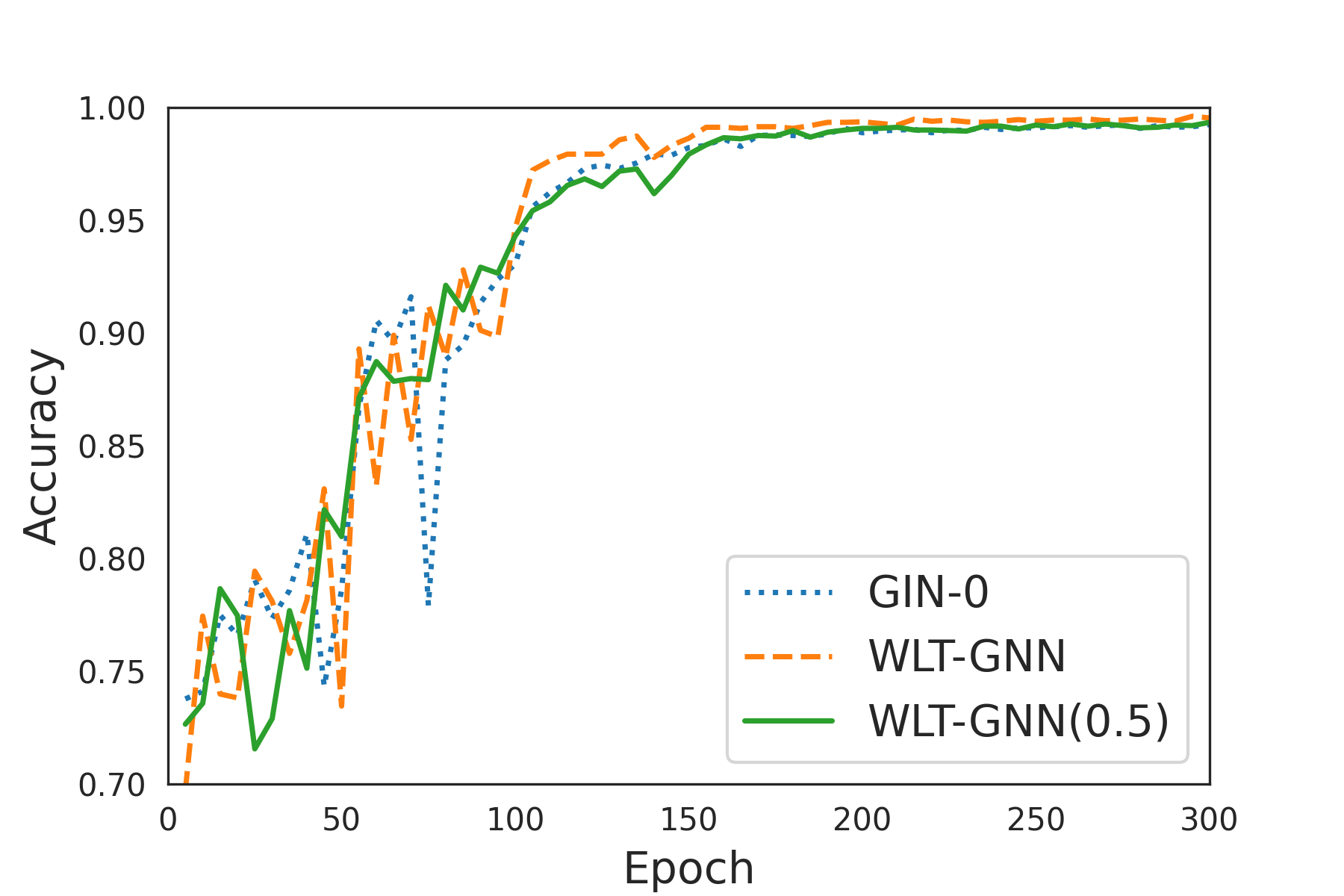}\\
                \includegraphics[width=1\textwidth,trim=0 0 30 30,clip]{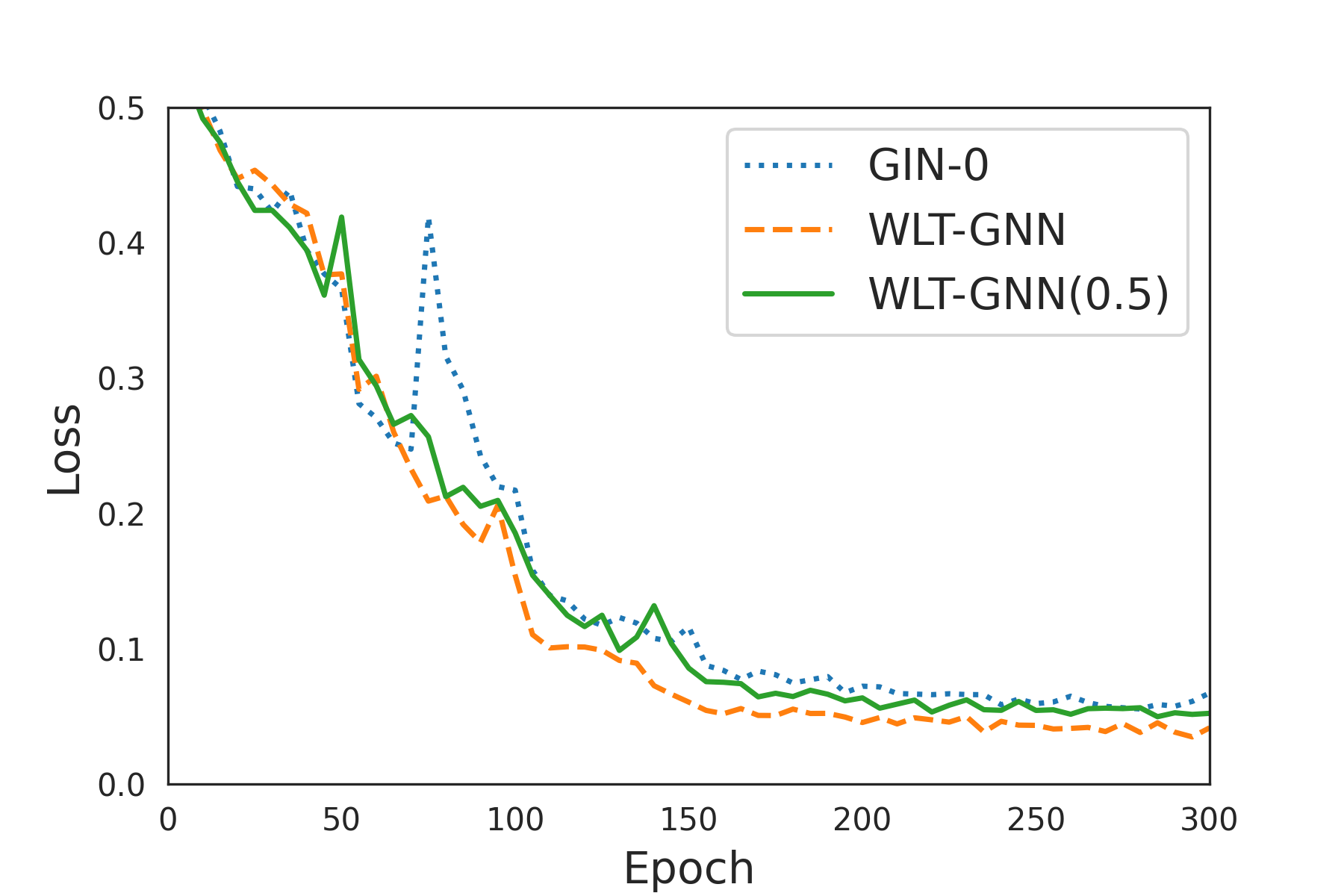}
            \end{minipage}
        }
        \subfigure[REDDIT-BINARY]{
            \begin{minipage}[b]{0.27\textwidth}
                \includegraphics[width=1\textwidth,trim=0 0 30 30,clip]{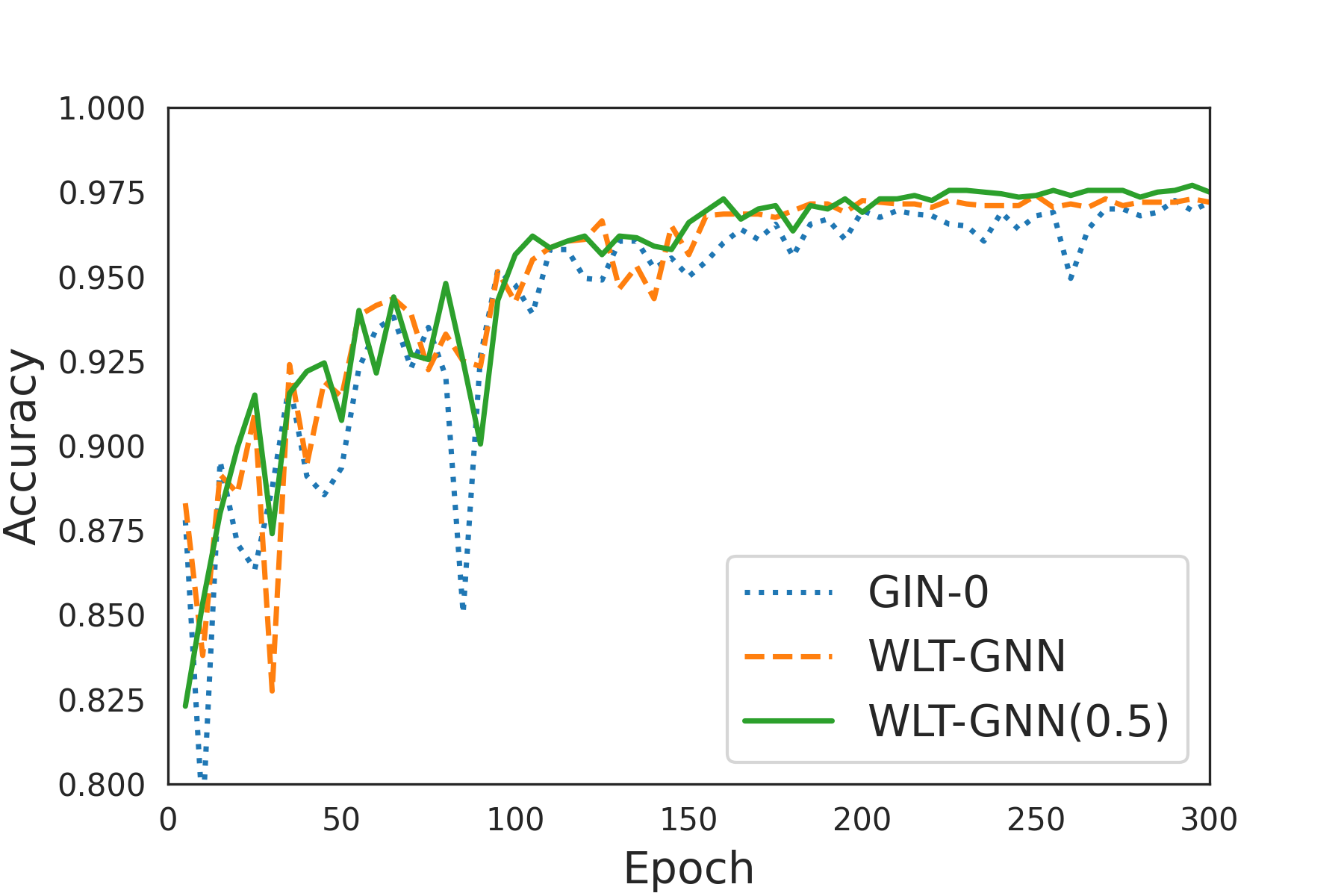}\\
                \includegraphics[width=1\textwidth,trim=0 0 30 30,clip]{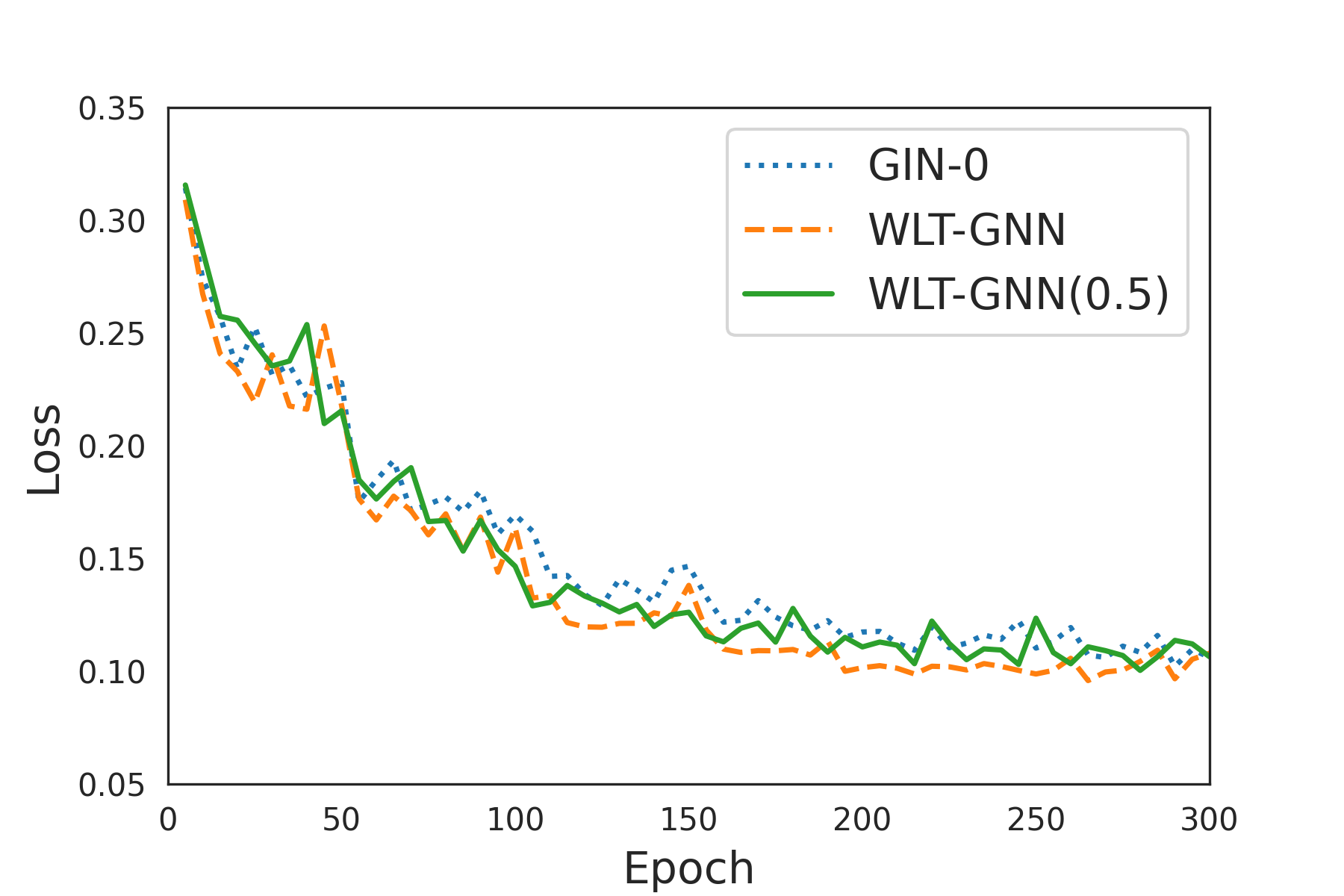}
            \end{minipage}
        }
    }
    \caption{The training accuracies and losses on the three datasets with respect to epochs.}
    \label{fig:training-acc}
    \end{center}
    \vskip -0.2in
    \end{figure*}

The results are reported in Table~\ref{train_result}. 
On all the three datasets, the WLT-GNN and WLT-GNN($0.5$) have better training accuracies compared with 
the GIN-$0$. 
In particular, the WLT-GNN($0.5$) reduces almost $50\%$ of the misclassified training graphs on PROTEINS. 
The training accuracies and losses with respect to epochs in the training procedure is plotted in Figure~\ref{fig:training-acc}. 
The curves show that on these three datasets, 
the proposed WLT-GNN and WLT-GNN($0.5$) have better training performance in terms of accuracy and fitting loss. 
We may infer that the recoloring layer helps improving the expressive power of GNN. 

\begin{table}
    \caption{\textbf{Training Accuracies ($\%$).} 
    The accuracies are the best in $5$ runs of training. 
    The best accuracies are highlighted with boldface.}
    \label{train_result}
    \begin{center}
    \begin{small}
    \begin{sc}
    \begin{tabular}{ccccc}
    Method         & PROTEINS       &NCI1   & REDDIT-B\\
    GIN-$0$        & 97.8 &  99.2   & 97.2        \\
    WLT-GNN        & 98.3 &  \textbf{99.5}   & 97.2 \\
    WLT-GNN($0.5$) & \textbf{99.0} &  99.3  & \textbf{97.5} \\
    \end{tabular}
    \end{sc}
    \end{small}
    \end{center}
    \vskip -0.1in
\end{table}

\section{Conclusion \& Future Works}
\label{conclusion}
In this paper, we proposed the WLT-GNN based on the message passing scheme and the Tinhofer algorithm. 
By introducing the recoloring layer to GNN, 
the expressive power of WLT-GNN can theoretically break through the limitation of WL algorithm. 
Further, we proposed the heuristic WLT-GNN($0.5$), which is assumed to work better on large graphs. 
In practice, we conducted comparative experiments to show that the WLT-GNN and WLT-GNN($0.5$) perform better on several 
well-known datasets. 
We also use training performance to show that the recoloring layer helps to improve the expressive power 
of GNN on three datasets. 

Introducing the recoloring layers enlarges the search scope of a good GNN. 
People may have different recoloring methods and arrangements of recoloring layers. 
Also, we are looking forward to mathematical proofs of generalized Tinhofer algorithms, for example, 
a Tinhofer algorithm compatible with multi-node recoloring operations.


\nocite{arvind2015tinhofer,arvind2017graph,tinhofer1986graph, tinhofer1991note,borgwardt2005protein,wale2006comparison,yanardag2015deep,morris2020tudataset,leskovec2005graphs}

\bibliography{alan-gnn}
\bibliographystyle{plain}


\appendix
\setcounter{equation}{0}
\setcounter{footnote}{0}
\renewcommand{\thefootnote}{\thesection.\arabic{footnote}}
\renewcommand{\theequation}{\thesection.\arabic{equation}}
\section{Compact Graph} \label{app:compact_graph}
In this part of the Appendix, we briefly introduce the compact graphs \cite{tinhofer1986graph, tinhofer1991note}. 

Let $G=(V(G), E(G))$ be a graph, where the $V(G), E(G)$ are the sets of vertices and edges, respectively. 
A permutation $\pi$ on the vertex set $V(G)$ is called an automorphism of $G$, 
if $\pi$ perserves the edges, \textsl{i.e., } for two vertices $u,v \in V(G)$, 
an edge $\{u,v\} \in E(G)$ iff $\{\pi(u),\pi(v)\}\in E(G)$. 
Let matrix $\bm{A}$ be the adjacent matrix of $G$. 
Using the permutation matrix $\bm{P}$ to represent a permutation $\pi$ on $V(G)$, then $\pi$ is an 
automorphism iff $\bm{P}$ commutes with the adjacent matrix $\bm{A}$, 
\begin{equation}\label{app:auto}
    \bm{P}\bm{A}=\bm{A}\bm{P}.
\end{equation}
In the following, we use $\mathrm{Aut}(\bm{A})$ to represent the solution set of Equation (\ref{app:auto}). 
A doubly stochastic matrix $\bm{X}$ is a square matrix with non-negative entries 
and the sum of all entries in any row or column is equal to $1$, 
mathematically, a doubly stochastic matrix $\bm{X}$ satisfies
\begin{equation}\label{app:dsm}
    \bm{X}\bm{e} = \bm{X}^t\bm{e} = \bm{e}, \quad \bm{X}\geq 0, 
\end{equation}
where $\bm{e}$ is a vector of $1$s. 
If we replace the permutation matrix $\bm{P}$, which is also a doubly stochastic matrix, 
in Equation (\ref{app:auto}) with a doubly stochastic matrix $\bm{X}$, 
\begin{equation}\label{app:dsm-auto}
    \bm{X}\bm{A}=\bm{A}\bm{X}, 
\end{equation}
the solutions of Equations (\ref{app:dsm}, \ref{app:dsm-auto}) form a subpolytope of $S_{|V(G)|}$. 
Let us use $S(\bm{A})$ to denote the solutions of Equations (\ref{app:dsm}, \ref{app:dsm-auto}) in the following. 

Using these notations, the compact graph is defined as
\begin{defination}
A graph $G$ with adjacent matrix $\bm{A}$ is called \textsl{compact} iff it satisfies the following condition:\\
Every doubly stochastic matrix $\bm{X}$ which commutes with $\bm{A}$ 
is a convex sum of automorphisms of $\bm{A}$.
\end{defination}
In detail, a graph is compact iff for any $\tilde{\bm{X}} \in S(\bm{A})$, 
\begin{equation}
    \tilde{\bm{X}} = \sum_{\bm{P}_i\in \mathrm{Aut}(\bm{A})} a_i \bm{P}_i,
\end{equation}
where $a_i\geq 0$ and $\sum_i a_i = 1$. 
It is known that many kinds of graphs, to name a few, complete graphs, cycles, trees, \textsl{etc.}, are 
compact. 
It is also known that a graph that can be distinguished from any non-isomorphic graph by WL test is compact. 
More compact graphs and the relation between compactness, graph isomorphism, and WL test 
can be found in \cite{arvind2015tinhofer,arvind2017graph}.

\section{Datasets}\label{app:dataset}
The datasets used in this paper are two bioinformatic datasets \cite{borgwardt2005protein,wale2006comparison}: 
PROTEINS and NCI1, 
and five social network datasets \cite{yanardag2015deep}: 
COLLAB, IMDB-BINARY, IMDB-MULTI, REDDIT-BINARY, and REDDIT-MULTI-5K. 
In the experiments, these datasets are obtained through the Python package 
\textsl{TUDataset}\footnote{https://github.com/chrsmrrs/tudataset} \cite{morris2020tudataset}. 

PROTEINS is a dataset whose samples are graphs representing proteins. 
In each graph, the nodes represent the secondary structure elements and have labels of helix, sheet, or turn. 
If two nodes are neighbors along the amino acid sequence or in 3D space, there is an undirected edge connecting them. 

NCI1 is a balanced dataset of chemical compounds screened for activity against non-small cell lung cancer. 
Each graph in NCI1 represents a chemical compound, where the nodes, node labels, edges are related to the 
atoms, atom types, and chemical bonds, respectively. 

COLLAB is a scientific collaboration dataset. It is derived from 
three scientific collaboration datasets \cite{leskovec2005graphs}, namely, 
High Energy Physics, Condensed Matter Physics, and Astro Physics. 
The graphs are the ego-networks of different researchers from each field. 

IMDB-BINARY and IMDB-MULTI are datasets of movie collaborations. They contain ego-networks 
derived from each actor/actress by the collaboration relations. 
The labels of the graphs in IMDB-BINARY are genres of Action and Romance, 
while the IMDB-MULTI have graph labels according to genres of Comedy, Romance, and Sci-Fi. 

REDDIT-BINARY and REDDIT-MULTI-5K are balanced datasets similar to the IMDB-BINARY and IMDB-MULTI. 
Each graph in REDDIT datasets represents an online discussion thread by representing the users as nodes in the graph. 
Two users are connected by an undirected edge if anyone responded to another's comment. 
The graph labels of REDDIT-BINARY are discussion or question/answer, 
while the graphs in REDDIT-MULTI-5K have labels according to their subreddits.

\end{document}